\documentclass[a4paper]{article}
\usepackage[dvipdfmx]{graphicx}
\usepackage{here}
\usepackage{paralist}
\usepackage{multirow}
\usepackage[hidelinks]{hyperref}
\usepackage{color, soul}

\usepackage{INTERSPEECH2022}

\title{Deep versus Wide: An Analysis of Student Architectures\\for Task-Agnostic Knowledge Distillation of Self-Supervised Speech Models}
\name{Takanori Ashihara, Takafumi Moriya, Kohei Matsuura, Tomohiro Tanaka}
\address{NTT Corporation, Japan}
\email{takanori.ashihara.vk@hco.ntt.co.jp}

\begin{document}

\maketitle
\begin{abstract}
\vspace{-0.1cm}
Self-supervised learning (SSL) is seen as a very promising approach with high performance for several speech downstream tasks.
Since the parameters of SSL models are generally so large that training and inference require a lot of memory and computational cost, it is desirable to produce compact SSL models without a significant performance degradation by applying compression methods such as knowledge distillation (KD).
Although the KD approach is able to shrink the depth and/or width of SSL model structures, there has been little research on how varying the depth and width impacts the internal representation of the small-footprint model.
This paper provides an empirical study that addresses the question.
We investigate the performance on SUPERB while varying the structure and KD methods so as to keep the number of parameters constant; this allows us to analyze the contribution of the representation introduced by varying the model architecture.
Experiments demonstrate that a certain depth is essential for solving content-oriented tasks (e.g.~automatic speech recognition) accurately, whereas a certain width is necessary for achieving high performance on several speaker-oriented tasks (e.g.~speaker identification).
Based on these observations, we identify, for SUPERB, a more compressed model with better performance than previous studies.
\end{abstract}

\vspace{-0.2cm}
\section{Introduction}
\label{intro}
\vspace{-0.1cm}
Self-supervised learning (SSL) has become the key technique not only for natural language processing (NLP) and computer vision (CV) communities, but also for the speech community.
This approach offers a general-purpose representation learned in an unsupervised fashion and achieves state-of-the-art performance in many downstream tasks.
Particularly in speech representation learning, many SSL studies have been published \cite{cpc, apc, wav2vec, pase, tera, mockingjay, decoar2, w2v2, hubert, wavlm, data2vec}.
They detail impressive performance and capability in solving the multiple tasks associated with the Speech processing Universal PERformance Benchmark (SUPERB) \cite{superb}.
While SSL is successful, the pre-trained models generally have high computation costs for training and inference due to huge over-parameterization, so they cannot be deployed on mobile devices with limited resources.
\par
To tackle the above problem, knowledge distillation (KD) \cite{kd}, which is a model compression method, has been used to transfer the knowledge from a large speech SSL model into a small, green model by following a teacher-student framework \cite{shrinkw2v2, distilt, distilhubert}.
There are two main approaches for KD with SSL models: task-specifically distillation and task-agnostic distillation.
The former is based on distilling the knowledge of a teacher model that has been fine-tuned for a specific-task.
For example, in \cite{shrinkw2v2, distilt}, an SSL model fine-tuned for the automatic speech recognition (ASR) task has been utilized to reduce the parameters without significant ASR performance degradation.
On the other hand, in task-agnostic distillation, an internal representation of a pre-trained teacher model is distilled to provide a universal small student model such as DistillHuBERT \cite{distilhubert}.
We focus on the latter approach in this paper because our aim is to benefit from its ability to solve multiple speech tasks with the same feature extraction module \cite{superb}.
In addition, task-specific distillation requires the preparation of teacher models fine-tuned to each task and so is more costly than task-agnostic distillation.
\par
When applying KD on an internal representation of itself, since the structure of deep neural networks (DNN) has the core property of accepting scaling with respect to its depth and width, the student model can be thinner and shallower than the teacher model which reduces model size.
However, there has been little understanding on how varying the depth and width of the small-footprint SSL model affects the internal speech representation and performance in downstream tasks.
In other words, do models with different structures distilled from the same teacher learn similar representations and achieve comparable performance?
Investigating this fundamental issue could form the basis for developing more resource-efficient networks such as EfficientNet \cite{efficientnet} in CV.
\par
In this paper, we empirically explore the question how to shrink student models in terms of depth and width. 
First, we use SUPERB to examine the performance realized while varying the depth and width of the student networks distilled from HuBERT \texttt{BASE}; the total number of parameters is kept the same as much as possible.
According to the result, we find the performance tendency that the larger the model size, the better the performance as reported in \cite{scaling}, and furthermore, that the deep\&narrow student is better than the shallow\&wide one in content-oriented tasks and vice versa in several of speaker-oriented tasks.
We also confirm that setting KD between student's last and teacher's intermediate layers such as DistilHuBERT is suitable for wide networks, whereas KD between intermediate layers is suitable for deep networks such as \cite{fitnets, moriya_distil}.
Additionally, we show a similar trend in the student model distilled from the HuBERT \texttt{LARGE} model.
This analysis contributes to practical requirements of speech applications because student model size remains the same even if the teacher size is increased.
Beyond this analysis, the smaller model that has an intermediate number of layers shows better performance than the previous method through the linear interpolation of prediction-layer and intermediate-layer losses.

\vspace{-0.2cm}
\section{Related Work}
\label{related}
\vspace{-0.1cm}
Distilling the representation from large to small task-agnostic models has attracted much attention in the NLP community, especially.
DistilBERT \cite{distilbert} set soft target distillation and cosine similarity losses between last layers.
Other works such as MobileBERT \cite{mobilebert} and TinyBERT \cite{tinybert} transferred knowledge by mapping intermediate layers.
More recently, MiniLM \cite{minilm, minilmv2} was proposed to attain a shallower architecture by carefully-designed self-attention KD.
Unfortunately, to the best of our knowledge, only DistilHuBERT has been proposed as a task-agnostic KD of speech representation.
These works focused on developing new effective KD methods, and little attention was paid to the effects of structures on the representations of students, especially for speech representation.
\par
Some previous works have offered analyses of SSL representations \cite{sim_anal, anal-self-att, prob_pr, layer-wise-anal, synth_anal, espnet_anal}.
For example, \cite{sim_anal} provided a similarity analysis between last-layer output of SSL methods/architectures.
In \cite{layer-wise-anal, synth_anal, espnet_anal}, the contribution of features was measured for various probing tasks across layers.
Their analysis provided sophisticated insights, but not much was elucidated about how the variations of fundamental network architecture yielded affected outputs.
\par
There have been several empirical studies related to deep versus wide network structures.
In the context of speech, shallow networks achieve similar performance to deep networks when mimicking the teacher's posterior in phoneme recognition \cite{pr_wide}, but \cite{dvsws} has shown that representation differences based on the number of layers seems to be related to robustness against noise.
In the CV community, deep structures trained by KD between intermediate-layers are crucial for performance in image classification tasks \cite{fitnets}, and \cite{cv_wad} has recently found that deep and wide models yield different characteristic error patterns but similar recognition accuracies.
Our study is inspired by these results as regards the distinctive representation between deep and wide architectures, and hence, we attempt to unveil the impact of varying the fundamental network structures (i.e.~depth and width) on speech representation/performance.
\par
Other work related to this paper includes \cite{scaling}, which investigated the relationship between the number of parameters and training loss of Mockingjay \cite{mockingjay} as well as between the number of parameters and performance on speaker recognition and phoneme classification tasks.
They demonstrated positive correlations between model size and loss/performance and exhibited power-law scaling nature on loss.
We follow this work by investigating the relationship between the variation of core model architecture and its performance.
In addition, we also focus on KD as a more efficient and practical way to train tiny models and also conduct more extensive experiments based on SUPERB which is one of the general benchmarks for speech representation.

\vspace{-0.5cm}
\section{Method}
\label{method}
\vspace{-0.2cm}
In this paper, for the purpose of analyzing how the representations of task-agnostic KD models vary with depth and width empirically, we apply two simple KD approaches: prediction-layer distillation and layer-to-layer (L2L) distillation methods.
While the former is based on DistilHuBERT \cite{distilhubert}, the latter is based on representation mapping layer-to-layer such as FitNets \cite{fitnets} to alleviate the difficulty of training deeper networks.
\par
For the whole network architecture, the teacher and student models are composed of 7-layers of convolutional neural networks (CNNs) with input of a raw-waveform, followed by multiple self-attention layers, as is commonly used in SSL studies.
For student models, we alter the depth and width of self-attention layers only while fixing the size of the CNNs.

\vspace{-0.3cm}
\subsection{Prediction-layer distillation}
\begin{figure}[ht]
  \centering
  \includegraphics[width=6cm]{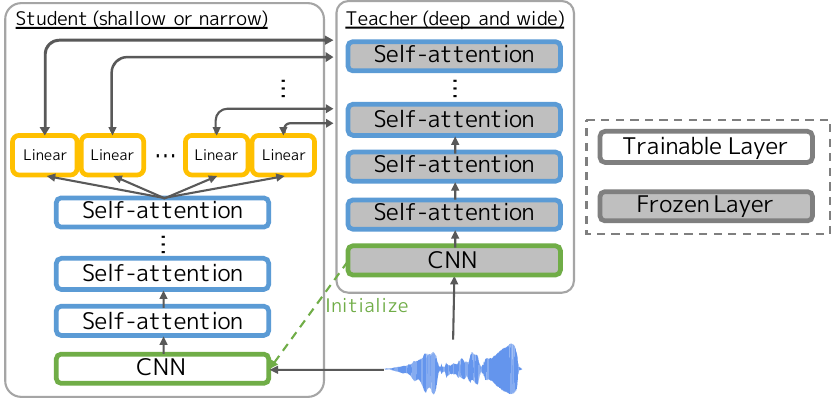}
  \caption{Illustration of student model trained by KD between student's last and teacher's intermediate layers based on DistilHuBERT \cite{distilhubert}.}
  \label{fig:method_last_layer}
\end{figure}

\vspace{-0.1cm}
As illustrated in Figure~\ref{fig:method_last_layer}, prediction-layer distillation transfers the representation by mapping between the last-layer of student and intermediate-layers of teacher models.
Since the effective teacher's representations for each downstream task are distributed layer-wise, DistilHuBERT \cite{distilhubert} utilizes the hidden states of empirically-selected multiple self-attention layers as teacher knowledge, and attains better performance than KD by utilizing only the last-layer of teacher models.
To project into the same dimensional space collected from multiple hidden states from the teacher, the prediction heads are added on the top layer of the student model and hence, the training objective is multi-task learning.
When we evaluate the student models, the heads are removed so as to follow DistilHuBERT.
In contrast to the original paper, we do not copy the parameters of the teacher model into the student model for initialization, except for convolutional layers, because we investigate not only shallow but also narrow models in which the hidden dimension size differs from those of larger teacher models.

\vspace{-0.3cm}
\vspace{-0.7cm}

\subsection{Layer-to-layer (L2L) distillation}
\begin{figure}[ht]
  \centering
  \includegraphics[width=6cm]{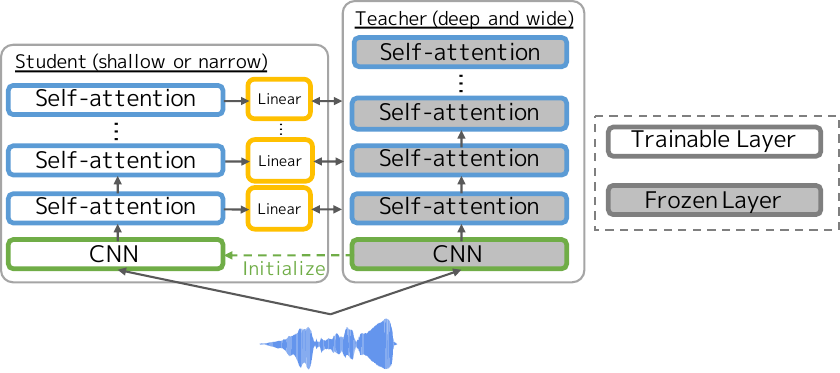}
  \caption{Illustration of student model trained by KD between intermediate-layers such as FitNets \cite{fitnets}.}
  \label{fig:method_inter_layer}
\end{figure}

\vspace{-0.1cm}
To train the deeper lightweight model efficiently, we also perform simple L2L distillation in addition to the former distillation as shown in Figure~\ref{fig:method_inter_layer}.
This approach, inspired by Fitnets, chooses the intermediate-layers from the teacher model as in DistilHuBERT, but transfers the knowledge to the student model by layer-to-layer mapping.
In the case that the intermediate layers from the narrow student yield smaller dimensional output than the teacher model, L2L KD also connects the projection layers to each intermediate layer.
The projection layers are removed when evaluation is conducted.
In L2L KD, we initialize only the student CNN block from the teacher model in the same way as the former KD and map the student to the teacher self-attention layers.

\vspace{-0.3cm}

\vspace{-0.6cm}
\subsection{Distillation objective}
\vspace{-0.2cm}
In this paper, as the objective function, we employ a combination of maximizing the L1 loss and minimizing the cosine similarity loss explored in DistilHuBERT.
The contribution of cosine similarity can be controlled by the weight factor (Equation (1) in \cite{distilhubert}), but we set their contribution equal in all experiments.

\begin{table*}[ht]
\caption{Evaluation result for each model distilled from HuBERT \texttt{BASE} and each task on SUPERB. The values in the first and second row are taken from \cite{superb} and \cite{distilhubert}, respectively. Pred. means the predication-layer distillation and L2L indicates the layer-to-layer distillation in the second column. For clarity, the KD models are indexed from (a) to (h) as shown in the second column.}
\vspace{-0.6cm}
\label{tab:result_base}
\begin{center}
\scalebox{0.88}[0.88]{
\begin{tabular}{l|l|c|c|c|c|c|c|c|c|c|c|c}
\toprule
\multirow{2}{*}{Model}      & \multirow{2}{*}{KD Loss} & PR    & ASR (w/ LM)   & KS    & QbE    & SID   & ASV  & SD   & IC    & SF            & ER    &      \\ \cline{3-12}
                           &            & PER$\downarrow$  & WER$\downarrow$   & Acc$\uparrow$   & MTWV$\uparrow$   & Acc$\uparrow$   & EER$\downarrow$  & DER$\downarrow$  & Acc$\uparrow$   & F1$\uparrow$ / CER$\downarrow$      & Acc$\uparrow$   & Rank$\downarrow$ \\ \hline
\midrule
HuBERT \texttt{BASE}       & \multicolumn{1}{|c|}{-} & 5.41  & 6.42 (4.79)  & 96.30 & 0.0736 & 81.42 & 5.11 & 5.88 & 98.34 & 88.53 / 25.20 & 64.92 & 1.7  \\
\midrule
DistilHuBERT                & \multicolumn{1}{|c|}{Pred.} & 16.27 & 13.34 (9.21) & 95.98 & 0.0511 & 73.54 & 8.55 & 6.19 & 94.99 & 82.57 / 35.59 & 63.02 & 5.8  \\ 
\midrule
\multirow{2}{*}{12-L \texttt{HALF}}   &  (a) Pred.                    & 13.09 & 11.87 (8.07) & 96.97 & 0.0501 & 69.11 & 6.32 & 6.67 & 94.91 & 84.49 / 32.54 & 62.76 & 4.6  \\
                            & (b) L2L                      & 10.67 & 10.96 (7.68) & 97.24 & 0.0604 & 69.52 & 6.13 & 6.81 & 96.97 & 86.11 / 30.93 & 63.24 &  2.6  \\ \hline
\multirow{2}{*}{12-L \texttt{FOURTH}} & (c) Pred.                     & 18.92 & 14.02 (9.25) & 96.44 & 0.0495 & 49.51 & 6.74 & 7.12 & 87.03 & 81.21 / 37.27 & 62.82 & 8.1  \\
                            & (d) L2L                      & 16.96 & 13.84 (9.20) & 96.40 & 0.0562 & 47.67 & 6.41 & 7.12 & 91.62 & 84.81 / 32.77 & 61.84 & 7.0  \\ 
\midrule
\multirow{2}{*}{3-L \texttt{ONE}}     & (e) Pred.                     & 13.34 & 12.23 (8.64) & 96.69 & 0.0489 & 75.71 & 6.48 & 6.56 & 94.15 & 82.89 / 34.65 & 63.95 & 4.6  \\
                            & (f) L2L                      & 13.96 & 12.94 (9.11) & 96.52 & 0.0568 & 47.76 & 6.18 & 7.17 & 96.02 & 85.99 / 32.38 & 62.57 & 5.2  \\ \hline
\multirow{2}{*}{3-L \texttt{HALF}}    & (g) Pred.                     & 18.62 & 13.91 (9.27) & 96.22 & 0.0482 & 62.59 & 6.86 & 6.69 & 91.88 & 82.78 / 35.75 & 61.83 & 8.1  \\
                            & (h) L2L                      & 18.11 & 14.48 (9.86) & 96.48 & 0.0502 & 60.40 & 6.82 & 7.31 & 94.91 & 81.82 / 37.36 & 62.78 &  7.5  \\
\bottomrule
\end{tabular}
}
\vspace{-0.8cm}
\end{center}
\end{table*}

\vspace{-0.35cm}
\section{Experimental Setup}
The teacher and student models explored in this paper are summarized in Table~\ref{tab:models}.
As teachers, we utilize HuBERT \texttt{BASE} and \texttt{LARGE}, both of which are publicly available from fairseq\footnote{https://github.com/pytorch/fairseq} \cite{fairseq}.
To build lightweight models, for the sake of simplicity, we reduced only the width of self-attention layers by two ratios, specifically one half and one fourth.
For the purpose of comparing the depth and width while keeping the total number of parameters constant, 12-L \texttt{HALF} and 12-L \texttt{FOURTH} were trained to compare 3-L \texttt{ONE} and 3-L \texttt{HALF}, respectively, which are comparable in terms of the number of parameters.
Note that when training the 3-L \texttt{ONE} model with L2L KD, we did not add the projection layer because the teacher and student self-attention layers had the same dimension originally.
6-L \texttt{HALF} is utilized for further analysis as discussed later.
\par
All student models were trained using the same settings as DistilHuBERT on a single GPU.
Specifically, the training steps totaled 200k with 24 batch size, and the models were optimized with Adam; the learning rate was linearly increased to 2e-4 for the first 7\% of updates, and then linearly decayed.
As the training dataset, we employed the standard Librispeech 960h \cite{librispeech} without labels.
When applying KD, the student models are mapped with the teacher representation by taking all layers for 12-L \texttt{HALF} and \texttt{FOURTH}, by taking 4th, 8th and 12th layers for 3-L \texttt{ONE} and \texttt{HALF} and by taking one layer out of two for 6-L \texttt{HALF}.

\begin{table}[h]
\caption{Model settings of teacher and student models. With respect to self-attention blocks, \texttt{HALF} and \texttt{FOURTH} means the parameter reductions to one half and one fourth, respectively, and \texttt{ONE} is the same as the HuBERT \texttt{BASE}.}
\vspace{-0.6cm}
\label{tab:models}
\begin{center}
\scalebox{0.80}[0.80]{
\begin{tabular}{l|ccccc}
\toprule
Models & \#Params & \#Layers & Embed. & FFN & \#Head \\   \hline
\midrule
HuBERT \texttt{BASE} \cite{hubert}  & 94.68M   & 12       & 768   & 3072       & 12    \\
HuBERT \texttt{LARGE} \cite{hubert}  & 316.61M   & 24       & 1024   & 4096       & 16    \\
\midrule
DistilHuBERT \cite{distilhubert} & 23.49M    & 2        & 768   & 3072       & 12    \\
\midrule
12-L \texttt{HALF}     &  26.87M   & 12       &  384  &  1536    &  6  \\
12-L \texttt{FOURTH}   & 9.93M    & 12       & 192  &  768    &  3  \\ 
\midrule
3-L \texttt{ONE}       &   30.58M  & 3        & 768  &  3072   & 12   \\
3-L \texttt{HALF}      &  10.90M  & 3        &  384  &  1536    &  6   \\
\midrule
6-L \texttt{HALF}      & 16.23M   & 6        &  384  &  1536  & 6  \\
\bottomrule
\end{tabular}
}
\vspace{-0.6cm}
\end{center}
\end{table}

\par
We evaluated the task-agnostic small KD models on SUPERB \cite{superb}.
SUPERB consists of 10 downstream tasks: phoneme recognition (PR), automatic speech recognition (ASR), keyword spotting (KS), query by example spoken term detection (QbE), speaker identification (SID), automatic speaker verification (ASV), speaker diarization (SD), intent classification (IC), slot filling (SF) and emotion recognition (ER).
These tasks are categorized into content, speaker, semantics and paralinguistics aspects as described in \cite{superb}.
This benchmark freezes the parameters of SSL models to be utilized as a feature extractor and weighted-sums the features collected from hidden layers except for QbE task.
The above training and evaluation processes were implemented with S3PRL\footnote{https://github.com/s3prl/s3prl} \cite{tera, mockingjay}.

\begin{table*}[ht]
\caption{Evaluation result for each model distilled from HuBERT \texttt{LARGE}. The values of first row are taken from \cite{superb}. The values shown from the second row are the results of the KD models trained in our experiment.}
\vspace{-0.65cm}
\label{tab:result_large}
\begin{center}
\scalebox{0.86}[0.86]{
\begin{tabular}{l|c|c|c|c|c|c|c|c|c|c|c|c}
\toprule
\multirow{2}{*}{Model}     & \multirow{2}{*}{KD Loss} & PR    & ASR (w/ LM)   & KS    & QbE    & SID   & ASV  & SD   & IC    & SF            & ER    &      \\ \cline{3-12}
                           &             & PER$\downarrow$  & WER$\downarrow$   & Acc$\uparrow$   & MTWV$\uparrow$   & Acc$\uparrow$   & EER$\downarrow$  & DER$\downarrow$  & Acc$\uparrow$   & F1$\uparrow$ / CER$\downarrow$      & Acc$\uparrow$   & Rank$\downarrow$ \\ \hline
\midrule
HuBERT \texttt{LARGE}               & \multicolumn{1}{|c|}{-} & 3.53  & 3.62 (2.94)  & 95.29 & 0.0353 & 90.33 & 5.98 & 5.75 & 98.76 & 89.81 / 21.76 & 67.62 &  2.7  \\
\midrule
\multirow{2}{*}{12-L \texttt{HALF}}  & Pred.                    & 9.67  & 9.59 (6.84)  & 95.79 & 0.0507 & 49.25 & 5.84 & 6.20 & 95.07 & 84.88 / 31.17 & 63.59 &  4.2  \\
                           & L2L                      & 7.97  & 9.24 (6.82)  & 96.24 & 0.0513 & 52.42 & 6.36 & 6.60 & 96.92 & 87.26 / 28.92 & 64.51 & 3.2  \\ \hline
\multirow{2}{*}{12-L \texttt{FOURTH}} & Pred.                & 14.10 & 12.49(8.47) & 96.20 & 0.0482 & 37.18 & 6.86 & 6.93 & 91.91 & 83.66/35.11 & 62.45 & 6.8 \\
			  & L2L                      & 12.86 & 12.91(9.11) & 95.34 & 0.0443 & 47.51 & 7.26 & 7.05 & 92.86 & 83.83/34.22 & 62.20 & 7.4 \\
\midrule
\multirow{2}{*}{3-L \texttt{ONE}}    & Pred.                     & 12.11 & 11.35 (8.00) & 96.50 & 0.0474 & 76.97 & 7.22 & 6.61 & 96.63 & 85.36 / 31.60 & 65.80 & 3.7 \\
                           & L2L                      & 10.24 & 12.23 (8.78) & 96.40 & 0.0540 & 68.90 & 7.59 & 7.33 & 96.97 & 84.56 / 32.88 & 65.22 & 4.3 \\ \hline
\multirow{2}{*}{3-L \texttt{HALF}}   & Pred.                     & 15.78 & 13.28(9.34) & 96.20 & 0.0430 & 60.17 & 7.17 & 6.77 & 94.02 & 84.67 / 33.82 & 64.55 & 5.9  \\
                           & L2L                      & 15.11 & 14.31 (9.84) & 96.01 & 0.0532 & 55.35 & 7.47 & 7.81 & 92.72 & 84.04 / 34.33 & 63.40 & 6.9  \\
\bottomrule
\end{tabular}
}
\end{center}
\vspace{-0.3cm}
\end{table*}

\begin{table*}[ht]
\vspace{-0.2cm}
\caption{Evaluation result for each model distilled from HuBERT \texttt{BASE}. The values in the fifth row represent the model trained by the linear interpolation loss (Comb.) between the prediction-layer and L2L losses.}
\vspace{-0.65cm}
\label{tab:result_interpolation}
\begin{center}
\scalebox{0.86}[0.86]{
\begin{tabular}{l|c|c|c|c|c|c|c|c|c|c|c|c}
\toprule
\multirow{2}{*}{Model}     & \multirow{2}{*}{KD Loss} & PR    & ASR (w/ LM)   & KS    & QbE    & SID   & ASV  & SD   & IC    & SF            & ER    &      \\ \cline{3-12}
                           &             & PER$\downarrow$  & WER$\downarrow$   & Acc$\uparrow$   & MTWV$\uparrow$   & Acc$\uparrow$   & EER$\downarrow$  & DER$\downarrow$  & Acc$\uparrow$   & F1$\uparrow$ / CER$\downarrow$      & Acc$\uparrow$   & Rank$\downarrow$ \\ \hline
\midrule
HuBERT \texttt{BASE}       & -      & 5.41  & 6.42 (4.79)  & 96.30 & 0.0736 & 81.42 & 5.11 & 5.88 & 98.34 & 88.53 / 25.20 & 64.92 & 1.3  \\
\midrule
DistilHuBERT                & Pred.                    & 16.27 & 13.34 (9.21)  & 95.98 & 0.0511 & 73.54 & 8.55 & 6.19 & 94.99 & 82.57 / 35.59 & 63.02 & 4.1  \\ 
\midrule
\multirow{3}{*}{6-L \texttt{HALF}} &  Pred.   & 15.14 & 12.72 (8.68) & 96.85 & 0.0504 & 67.06 & 6.36 & 6.81 & 93.75 & 83.65 / 34.35 & 63.72 & 3.4  \\
                         & L2L   & 13.40 & 12.66 (8.59) & 96.38 & 0.0545 & 62.90 & 6.85 & 6.95 & 95.86 & 83.80 / 33.51 & 63.09 &  3.3  \\
                         & Comb.  & 14.68 & 12.43 (8.51) & 96.77 & 0.0516 & 65.75 & 6.81 & 6.83 & 94.57 & 84.32 / 33.99   & 64.78 & 2.9   \\
\bottomrule
\end{tabular}
}
\vspace{-0.9cm}
\end{center}
\end{table*}

\vspace{-0.2cm}
\section{Results and Analysis}
\vspace{-0.1cm}
\subsection{How depth and width affects the performance}
\label{sec:exp1_1}
\vspace{-0.1cm}
Here, we analyze the relationship between the variation in student architectures with two KD methods and the resulting performances on SUPERB.
Table~\ref{tab:result_base} shows the KD results by using the knowledge of HuBERT \texttt{BASE}.
The ranking scores in the rightmost column indicate the relative values, which are the averages of each rank on the downstream tasks, as used in \cite{distilhubert}.
First of all, regarding the variation in KD loss function, prediction-layer loss is suitable for wider architectures such as (e) (the 7th row of the Table~\ref{tab:result_base}), whereas L2L loss is effective for deeper architectures such as (b) and (d).
By comparing (b) to (e) and (d) to (g, h), which have almost the same model size but different structures, we observe that while deeper networks have higher performance in content-oriented tasks such as PR, ASR and QbE, wider networks have higher performance in speaker-oriented tasks such as SID and SD.
Note that 3-L \texttt{ONE} and \texttt{HALF} were trained by imitating the representation suitable for the ASR task explicitly by heuristically selecting the layers whose representations are weighted for the ASR task (e.g.~8th layer on HuBERT \texttt{BASE}) \cite{distilhubert}.
From the above results, deep models seem to be effective when the task to be solved is based on contents while wide models support tasks related to speakers.
This is possible because the context size of self-attention is able to increase with the number of layers, and thus the deep architecture is advantageous for content-oriented tasks that require long context.
On the other hand, increasing the representation capacity within lower layers is seemingly effective for speaker-oriented tasks that focus local acoustic features rather than contextual information.
As shown in \cite{distilhubert}, it certainly demonstrates that effective features for SID (ASR) task are concentrated in the lower (upper) layers.
We also confirm a clear tradeoff between the number of parameters and overall performance which agrees with \cite{scaling, scaling_lm}.

\vspace{-0.2cm}
\subsection{Is the tendency the same for a larger teacher model?}
\vspace{-0.1cm}
From Section~\ref{sec:exp1_1}, the student models emphasize the characteristic performance in the case of HuBERT \texttt{BASE}, do the models distilled from HuBERT \texttt{LARGE} show a similar property?
These results are important for practical concerns because student size is unaffected by the teacher size.
Table~\ref{tab:result_large} shows the results, which are similar to Table~\ref{tab:result_base}, but different in that the teacher model is HuBERT \texttt{LARGE}, resulting students with better performance on PR, ASR and SF tasks in particular.
Note that the 12-L and 3-L models mimic the teacher representations of the even-numbered and multiple of eight layers, respectively.
Although we can find a performance tendency that follows that in Section~\ref{sec:exp1_1} on the PR, ASR and SID tasks, the students appear to be inconsistent with the former trends in some tasks such as QbE and KS.
However, identifying the reason is difficult because the performance deterioration is apparent in the original application of HuBERT \texttt{LARGE} compared with HuBERT \texttt{BASE} on these tasks, so HuBERT \texttt{LARGE} is considered to have learned a comparatively ineffective representation in such tasks.

\begin{table}[ht]
\caption{Result for each shallow model. Pred.-all and L2L-n-of-m are KD method of utilizing all HuBERT \texttt{BASE} layers.}
\vspace{-0.5cm}
\label{tab:result_nofm}
\begin{center}
\scalebox{0.91}[0.91]{
\begin{tabular}{l|c|c|c|c|c}
\toprule

\multirow{2}{*}{Model}   & \multirow{2}{*}{KD Loss} & PR    & ASR   & SID & IC    \\ \cline{3-6} 
                         &           & PER$\downarrow$   & WER$\downarrow$   & Acc$\uparrow$ & Acc$\uparrow$    \\ \hline
\midrule
\multirow{4}{*}{3-L \texttt{ONE}}  & Pred.                    & 13.34 & 12.23 & 75.71 & 94.15 \\
                         & Pred.-all                & 13.85 & 12.73 & 71.37   & 94.33 \\ \cline{2-6} 
                         & L2L                      & 13.96 & 12.94 & 47.76 & 96.02 \\
                         & L2L-n-of-m                 & 14.64 & 13.41 & 44.56   & 95.57 \\
\midrule
\multirow{4}{*}{3-L \texttt{HALF}} & Pred.                    & 18.62 & 13.91 & 62.59 & 91.88 \\ 
                         & Pred.-all                & 18.68 & 14.18 & 59.73   & 92.17 \\ \cline{2-6} 
                         & L2L                      & 18.11 & 14.48 & 60.40 & 94.91 \\
                         & L2L-n-of-m                 & 18.30 & 14.46 & 36.57   & 94.44 \\
\bottomrule
\end{tabular}
}
\end{center}
\vspace{-0.8cm}
\end{table}

\vspace{-0.1cm}
\subsection{Is it due to lack of information from the teacher?}
\vspace{-0.1cm}
The above results could be due to not capturing all teacher representations.
In other words, only selected layers are utilized for teacher prediction for KD.
Therefore, we adopted an additional training approach of utilizing all teacher layers.
Although DistilHuBERT makes it easy to scale the number of teachers layers by adding the prediction heads to student models (called Pred.-all), it is difficult to achieve with L2L KD.
To alleviate this problem, some attempts have been made for efficient intermediate layer mapping \cite{alp-kd, skip, rail-kd} in the NLP community; we adopt the simpler method inspired by the RAIL-KD \cite{rail-kd} for our analysis.
When applying L2L KD, this method randomly selects as many layers as the student model from all layers of the teacher model at each batch (referred to as L2L-n-of-m).
Table~\ref{tab:result_nofm} compares the performance on PR, ASR, SID and IC for two shallow models (3-L \texttt{ONE} and \texttt{HALF}), trained by different KD losses. 
The results show there is no significant improvement in performance for both methods.
Perhaps there is a need to simply advance the KD method which is better-suited for speech representation.
However, in our experiment, the severe degradation in PR/ASR performance seems not to be due to omission in teacher layer selection but to structures.

\vspace{-0.15cm}
\subsection{Can a student that has an intermediate number of layers offers universal performance?}
\vspace{-0.15cm}
The above results can be summarized as narrow or shallow models have advantages and disadvantages in terms of performance.
Therefore, we built a KD model consisting of an intermediate number of layers, specifically 6-layers.
Table~\ref{tab:result_interpolation} presents the performance of KD models that were transferred knowledge from HuBERT \texttt{BASE}.
To construct the more universal model, we prepared a combination model trained by the linear interpolation loss between prediction-layer and L2L distillation losses, which corresponds to the 5th row in Table~\ref{tab:result_interpolation}.
In this experiment, we decided to weight the former loss term by 0.8 and the latter term by 0.2 based on our preliminary experiment.
The result is that our model offers not only better performance but also greater compression than DistilHuBERT.
\vspace{-0.3cm}
\section{Conclusions}
\vspace{-0.1cm}
In this paper, we empirically investigated and analyzed how varying the depth and width of small network structures impacted the speech representation formed by task-agnostic KD from a large SSL model.
Our findings indicate that a deep\&narrow student is better than the shallow\&wide equivalent in content-oriented tasks, while the reverse is true for several speaker-oriented tasks.
We also developed and tested a smaller model with better performance than the previous approach.
In future work, we will investigate other benchmarks such as SUPERB-SG \cite{superb-sg} to confirm these findings.
\clearpage
\bibliographystyle{IEEEtran}

\bibliography{mybib}

\end{document}